\documentclass{article}
\PassOptionsToPackage{numbers, compress}{natbib}

\usepackage[final]{new_in_ML}

\usepackage[utf8]{inputenc} 
\usepackage[T1]{fontenc}    
\usepackage{hyperref}       
\usepackage{url}            
\usepackage{booktabs}       
\usepackage{amsfonts}       
\usepackage{nicefrac}       
\usepackage{microtype}      
\usepackage{xcolor}         
\usepackage{graphicx}
\usepackage{subcaption}
\usepackage{multirow}
\usepackage{balance}
\usepackage{amsmath}
\title{DARLEI: Deep Accelerated Reinforcement Learning with Evolutionary Intelligence}

\author{
Saeejith Nair$^{1}$ \quad  Mohammad Javad Shafiee$^{1,2}$ \quad Alexander Wong$^{1,2}$\\
$^1$University of Waterloo, Waterloo, Ontario, Canada\\
$^2$Waterloo Artificial Intelligence Institute, Waterloo, Ontario, Canada\\
{\tt\small {\{smnair, mjshafiee, a28wong\}}@uwaterloo.ca}
}

\begin{document}

\maketitle

\begin{abstract}
We present DARLEI, a framework that combines evolutionary algorithms with parallelized reinforcement learning for efficiently training and evolving populations of UNIMAL agents. Our approach utilizes Proximal Policy Optimization (PPO) for individual agent learning and pairs it with a tournament selection-based generational learning mechanism to foster morphological evolution. By building on Nvidia's Isaac Gym, DARLEI leverages GPU accelerated simulation to achieve over 20x speedup using just a single workstation, compared to previous work which required large distributed CPU clusters. We systematically characterize DARLEI's performance under various conditions, revealing factors impacting diversity of evolved morphologies. For example, by enabling inter-agent collisions within the simulator, we find that we can simulate some multi-agent interactions between the same morphology, and see how it influences individual agent capabilities and long-term evolutionary adaptation. While current results demonstrate limited diversity across generations, we hope to extend DARLEI in future work to include interactions between diverse morphologies in richer environments, and create a platform that allows for coevolving populations and investigating emergent behaviours in them. Our source code is also made publicly available\footnote{Project website:  \url{https://saeejithnair.github.io/darlei}}.
\end{abstract}

\section{Introduction}
The diversity and complexity of life on Earth are testaments to the evolutionary process's creative and adaptive capabilities. However, despite extensive research into evolutionary algorithms, modern implementations still fall short in capturing the open-ended creativity inherent in natural evolution.~\cite{soros_open-endedness_2017}. Traditional methods, such as genetic programming, are typically goal-oriented and focus on optimizing predefined solutions. This approach misses a crucial element: the unceasing inventiveness and adaptability that characterizes natural evolutionary processes. This gap can perhaps be bridged by exploring coevolutionary dynamics~\cite{soros_open-endedness_2017}, where populations evolve in response to each other to yield more open-ended and innovative evolutionary outcomes. 

Consider the approach of Minimal Criterion Coevolution (MCC)~\cite{brant_minimal_2017}, where evolving environments, such as mazes, coevolve with the agents navigating them. As the complexity of these environments increases, agents are compelled to adapt and develop more sophisticated navigation strategies. This reciprocal evolution drives the complexity further, illustrating the potential for open-ended evolution. However, current implementations of MCC, often constrained to simple 2D gridworlds, do not fully leverage the possibilities. To tap into the fuller potential of such coevolutionary dynamics, a more advanced simulation framework is needed. This framework should enable the procedural generation of realistic, physics-based environments, support the evolution of a wide range of embodied morphologies, ensure scalable and efficient execution, and facilitate complex multi-agent interactions to uncover emergent ecological and evolutionary dynamics.

Recent developments in simulation tools, such as Nvidia's Omniverse Isaac Gym
~\cite{makoviychuk_isaac_2021}, along with progress in sim2real transfer~\cite{kadian_sim2real_2020} techniques, have set the stage for the creation of such sophisticated simulation platforms. A notable example is the DERL framework~\cite{gupta_embodied_2021}, which pioneered a distributed system for the automated design and training of embodied agents, tackling complex locomotion and manipulation tasks. Despite its promising outcomes, DERL's reliance on distributed CPU clusters poses a significant barrier, limiting accessibility for a broad range of researchers.

To overcome these limitations, we introduce Deep Accelerated Reinforcement Learning with Evolutionary Intelligence (DARLEI), a framework that refines and extends the core concepts of DERL. DARLEI leverages the power of GPU-accelerated simulation through Isaac Gym to realize a significant speedup of over 20x compared to DERL, while requiring just a single workstation. While our efficiency gains have so far only been demonstrated on locomotion tasks across simple planes, we hope that DARLEI can set the stage for more advanced research into multi-agent interactions and coevolutions within richly simulated environments.

\section{Methods}

DARLEI enables large-scale evolutionary learning by combining a distributed asynchronous architecture with GPU-accelerated simulation. It builds upon the UNIMAL~\cite{gupta_embodied_2021} design space and tournament selection approach of DERL, while harnessing the parallelism and speed of Isaac Gym for agent training.

\begin{figure}[h]
\centering
\begin{subfigure}{0.24\textwidth}
\includegraphics[width=1.0\linewidth, height=5cm]{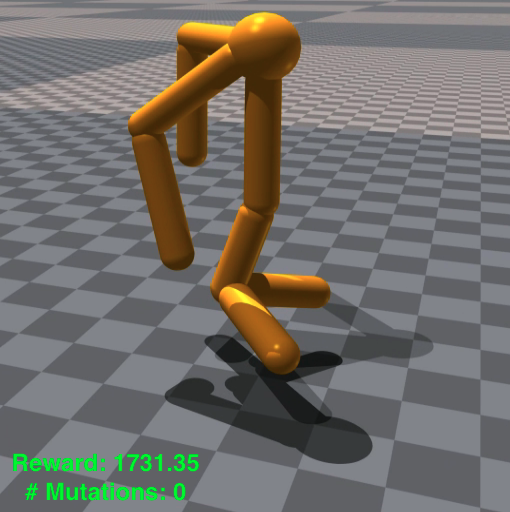} 
\label{fig:Mutation_example_1}
\end{subfigure}
\begin{subfigure}{0.24\textwidth}
\includegraphics[width=1.0\linewidth, height=5cm]{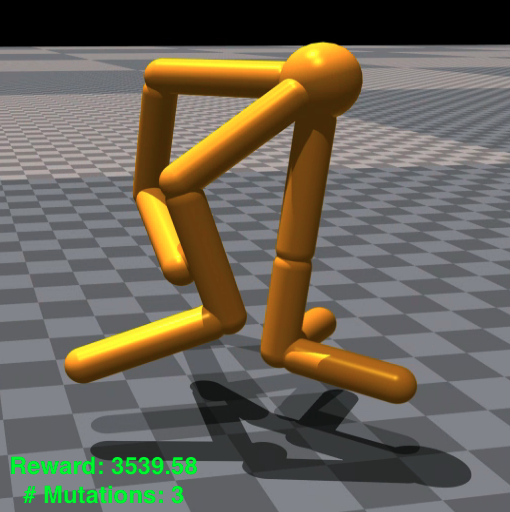}
\label{fig:Mutation_example_2}
\end{subfigure}
\begin{subfigure}{0.24\textwidth}
\includegraphics[width=1.0\linewidth, height=5cm]{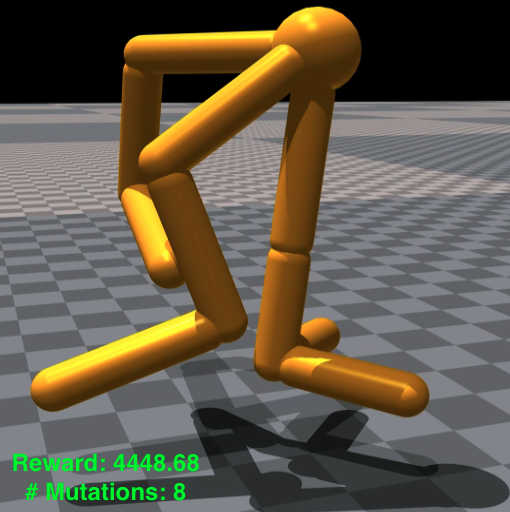}
\label{fig:Mutation_example_3}
\end{subfigure}
\begin{subfigure}{0.24\textwidth}
\includegraphics[width=1.0\linewidth, height=5cm]{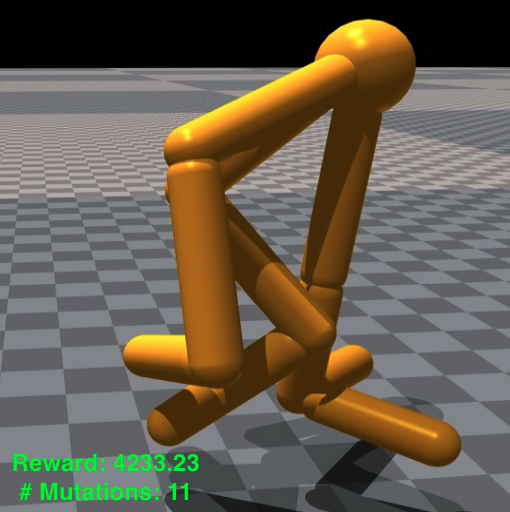}
\label{fig:Mutation_example_4}
\end{subfigure}

\caption{Evolution of the optimal agent in experiment with configuration P=100, T=50, W=20: This series illustrates morphological changes through successive mutations. While visually similar, significant but non-apparent adjustments in limb parameters, such as joint angles and density, occurred between mutations 3 and 8, enhancing the agent's fitness. Subsequent mutations, however, proved detrimental, leading to a decline in performance.}
\label{fig:mutation_examples}
\end{figure}

\subsection{System Architecture}

DARLEI employs a distributed asynchronous architecture similar to DERL, with separate worker processes for population initialization, agent training, and tournament evolution. This decouples the different stages, allowing them to be parallelized across CPU and GPU resources.

\begin{figure}[h]
    \centering
    \begin{subfigure}{0.3\textwidth}
        \centering
        \includegraphics[width=1.0\linewidth, height=4cm]{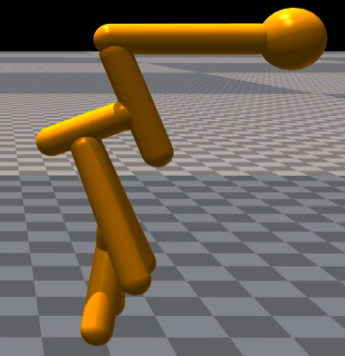} 
        \label{fig:Unimal_example_1}
    \end{subfigure}
    \hfill
    \begin{subfigure}{0.3\textwidth}
        \centering
        \includegraphics[width=1.0\linewidth, height=4cm]{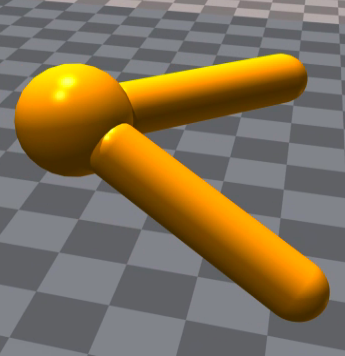}
        \label{fig:Unimal_example_2}
    \end{subfigure}
    \hfill
    \begin{subfigure}{0.3\textwidth}
        \centering
        \includegraphics[width=1.0\linewidth, height=4cm]{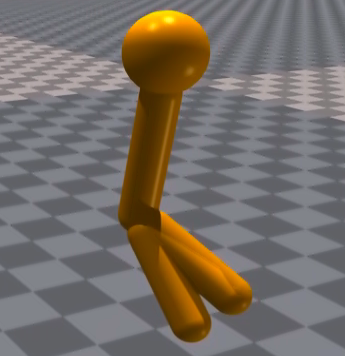}
        \label{fig:Unimal_example_3}
    \end{subfigure}
    \caption{Examples of agent morphologies from the initial population.}
    \label{fig:initial_population_morphologies}
\end{figure}

The core element borrowed from DERL is the UNIMAL (UNIversal aniMAL) design space, enabling the learning of locomotion and manipulation skills in stochastic environments without needing an accurate model of the agent or environment. UNIMAL agents are hierarchical rigid-body structures, generated procedurally through mutation operations starting from a root node. This genotype generation is conceptually similar to the morphological generation proposed in Evolved Virtual Creatures~\cite{sims_evolving_1994}, with the key distinction that agents are limited to 10 limbs and cyclic graphs are forbidden. Population initialization runs on the CPU, leveraging multiple processes to generate $P$ topologically unique UNIMALs from an initial pool of $10P$ candidate morphologies. Proprioceptive force sensors are then added to "foot" limbs before serializing to a MuJoCo-based XML representation~\cite{todorov_mujoco_2012} on a filesystem that all nodes and workers have access to. Figure~\ref{fig:initial_population_morphologies} shows examples of some valid morphologies that generated as part of the initial population.

\subsection{Agent Training}

In DARLEI, every UNIMAL agent undergoes a process called individual learning. This involves training through Proximal Policy Optimization (PPO)~\cite{schulman_proximal_2017} across 30 million simulation steps, aimed at learning locomotion tasks. These steps are parallelized across $M$ Isaac Gym environments on the GPU. We utilize Isaac Gym's default hyperparameters that were tuned for the Ant demo task. The training utilizes Isaac Gym's default hyperparameters, which were initially optimized for the Ant demo task. While there is potential for performance improvement or reduction in training steps through further hyperparameter tuning, our current experiments adhere to these default settings, leaving optimization for subsequent research.

\begin{figure}
  \centering
  \includegraphics[width=\linewidth]{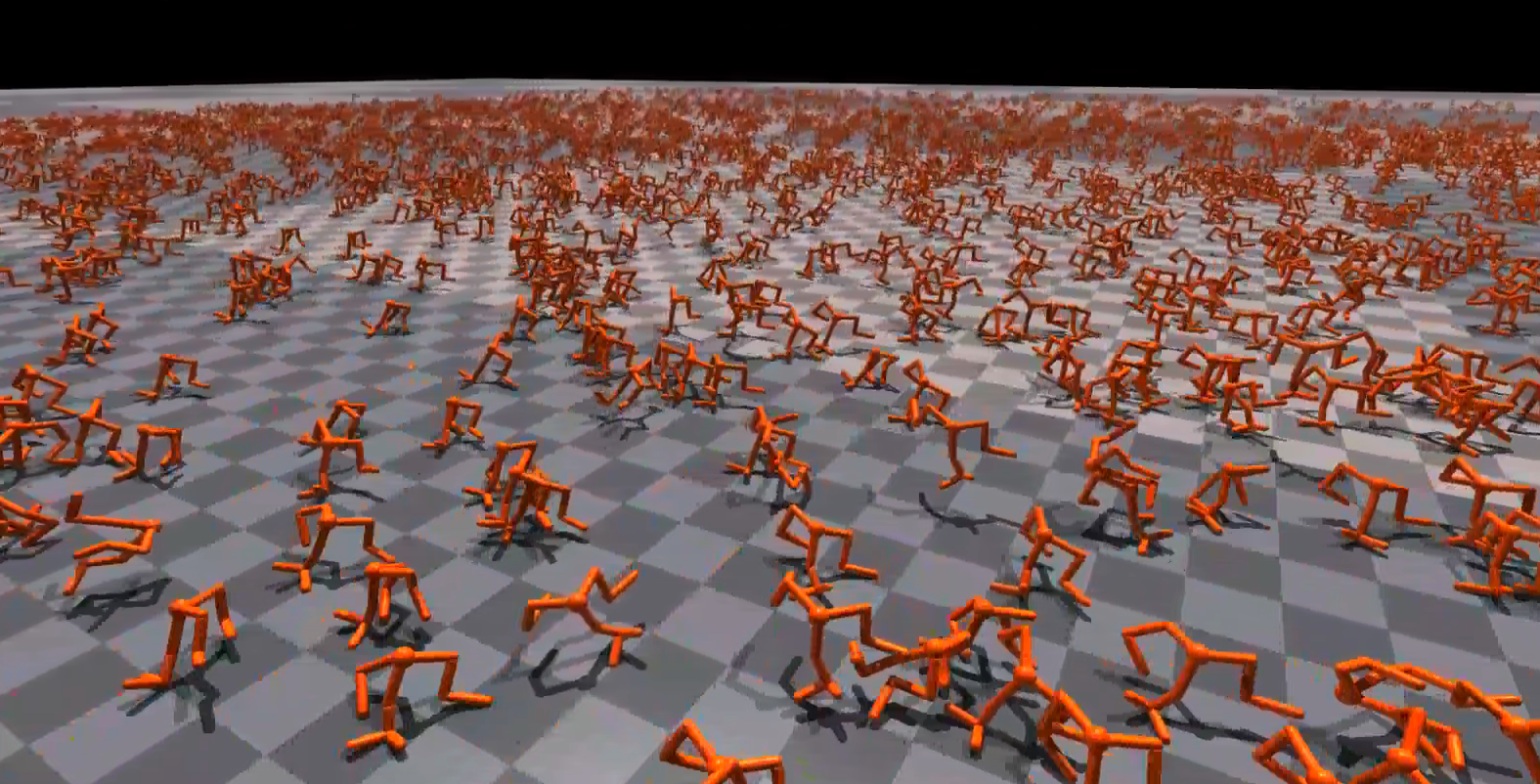}

  \caption{Overhead view of 8192 agents in Isaac Gym simulation.}
  \label{fig:simulation_parallel_env_overhead}
\end{figure}

Throughout the training phase, agents are limited to proprioceptive inputs, such as joint positions, velocities, and force sensor data, as well as ego-centric exteroceptive observations like head position and velocity relative to a target. Table \ref{tab:obs_space} shows a detailed list of the observation space. The primary evaluation scenario in our study is a simple environment, as illustrated in Figure~\ref{fig:simulation_parallel_env_overhead} where agents are tasked with moving towards a fixed target on flat terrain. Although integrating a variety of environments is feasible within our framework, our initial focus is on this specific flat terrain task, with plans to explore more diverse environments in future work.

\begin{align}
R = & \ R_{\text{progress}} + R_{\text{alive}} \times \mathbf{1}_{\{\text{head\_height} \ge \text{termination\_height}\}} \nonumber \\
    & + R_{\text{upright}} + R_{\text{heading}} \nonumber \\
    & + R_{\text{effort}} + R_{\text{act}} + R_{\text{dof}} \nonumber \\
    & + R_{\text{death}} \times \mathbf{1}_{\{\text{head\_height} \le \text{termination\_height}\}}
\label{eq:ft_reward}
\end{align}

The reward function (Equation \ref{eq:ft_reward}) is designed to encourage agents to move forward towards the target while maintaining an upright posture and avoiding early termination. This function closely aligns with those used in Isaac Gym's Ant and Humanoid demonstrations~\cite{makoviychuk_isaac_2021} but includes a significant modification: the termination height is dynamically set to half of the agent's initial head height. This adjustment, as suggested by DERL~\cite{gupta_embodied_2021}, aims to mitigate the tendency for excessive crawling behaviors. The effectiveness of the agent's learning (i.e. fitness) is quantified by calculating the average reward over the last 100,000 steps of its training period.

\subsection{Tournament Evolution}

Following the initial training phase, DARLEI starts the process of tournament evolution, executed asynchronously across $W$ parallel worker processes. In each iteration, workers independently sample 4 agents for competition. These agents are chosen uniformly at random from a pool spanning the range $[T \cdot G, Q]$, where $G$ represents the current generation number, $T$ is the tournaments held per generation, $P$ denotes the initial population size, and $Q$ is the cumulative count of evolved agents. The current generation number $G$ is calculated using the formula $G = \lfloor (Q-P)/T \rfloor$.

In each tournament, the four randomly selected agents are pitted against each other, with the agent exhibiting the highest fitness emerging as the victor. It’s important to note that the fitness values for each agent are determined once, during their initial phase of individual learning. The winning agent is then subjected to a mutation process, wherein a random modification from the UNIMAL design space is applied. This mutation could involve various alterations such as deleting or adding limbs, or changing limb characteristics like length, angle, and density. The newly mutated offspring is then reintegrated into the population, ready for participation in subsequent tournaments. This evolutionary loop is sustained for up to a maximum of 10 generations, a limitation set due to time constraints in our study.

In line with the strategies implemented in DERL, DARLEI utilizes an aging criterion to preserve population diversity and counteract the influence of initially fortunate genotypes. This criterion is based on a predefined range $R$. Similar to the parent queue in Chromaria~\cite{soros2014identifying}, aging in DARLEI serves as an egalitarian mechanism, ensuring that all agents, irrespective of their fitness levels, are eventually phased out due to age. This approach differs from direct elimination of low-fitness agents and offers a more balanced strategy. Furthermore, aging based on the number of completed generations, rather than the sheer population size, enhances the system's fault tolerance. In instances of worker failure, new workers can be seamlessly integrated without disrupting the existing population dynamics. As a result, our population size may temporarily exceed $P$ until the culmination of the ongoing generation. Operating with fewer workers than DERL, our approach allows for a greater number of mutations per agent by delaying aging until the completion of full generations, as opposed to prematurely aging them based on population size alone.

Our experimental setup involved a workstation equipped with dual NVIDIA A6000 GPUs and a 32-core AMD Ryzen Threadripper PRO 3955WX CPU. To aim for precise benchmarking, all experiments were conducted in isolation, ensuring no other applications were active during the testing period.

\begin{table}[]
\caption{Observation space used for training a UNIMAL. $\mathbb{A},\mathbb{F}$ refer to the number of actuators (joints), and feet respectively.}
\vspace{0.1in}
\label{tab:obs_space}
\centering
\begin{tabular}{l|l|l}
\hline
\multicolumn{2}{l|}{Observation Space}            & Degrees of Freedom                \\ \midrule \midrule
\multicolumn{2}{l|}{Head vertical position}      & 1                                 \\ \midrule
\multirow{2}{*}{Velocity}           & positional & 3                                 \\ 
                                    & angular    & 3                                 \\ \midrule
\multicolumn{2}{l|}{Yaw, roll, angle to target}   & 3                                 \\ \midrule
\multicolumn{2}{l|}{Up and heading vector proj.}  & 2                                 \\ \midrule
\multirow{2}{*}{DOF measurements}   & position   & $\mathbb{A}$                       \\ 
                                    & velocity   & $\mathbb{A}$                       \\ \midrule
\multicolumn{2}{l|}{Sensor forces}                & $\mathbb{F}$                            \\ \midrule
\multicolumn{2}{l|}{Sensor torques}               & $\mathbb{F}$                            \\ \midrule
\multicolumn{2}{l|}{Actions}                      & $\mathbb{A}$                       \\ \midrule \midrule
\multicolumn{2}{l|}{Total number of observations} & $12 + 3\mathbb{A}+ 2\mathbb{F}$ \\ \hline
\end{tabular}
\end{table}

\section{Results}
Our experiments with DARLEI assess its performance, scalability, and the quality of the solutions it evolved.

\subsection{Scalability via Parallel Environments}

\begin{figure}[h]
\centering
\begin{subfigure}{0.49\textwidth}
\includegraphics[width=\linewidth, height=7cm]{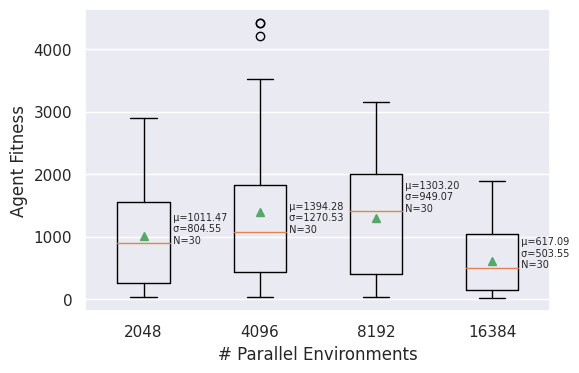} 
\label{fig:num_envs_vs_reward}
\end{subfigure}
\begin{subfigure}{0.49\textwidth}
\includegraphics[width=\linewidth, height=7cm]{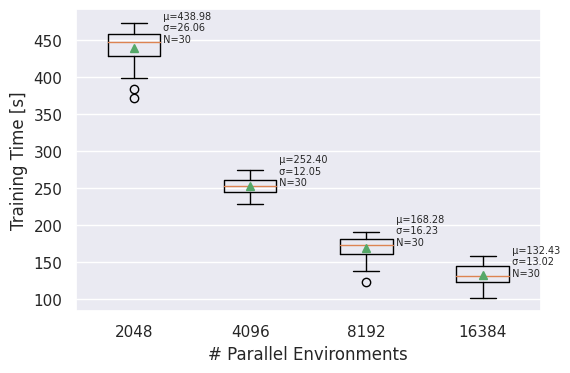}
\label{fig:num_envs_vs_train_time}
\end{subfigure}
\caption{Impact of parallel environments on agent fitness and training duration. To maintain consistency in the overall experience gained by each RL agent, the horizon length is proportionally reduced with an increase in the number of environments~\cite{makoviychuk_isaac_2021}. For example, we used horizon lengths of 64, 32, 16, and 8 for 2,048, 4,096, 8,192, and 16,384 environments, respectively. The data presented here are derived from the lifetime learning of 30 agents, all originating from the initial population set.}
\label{fig:num_envs}
\end{figure}

One of DARLEI's key strengths lies in its ability to employ a large number of parallel environments during training, significantly accelerating the process. Our findings, depicted in Figure~\ref{fig:num_envs}, show that increasing the number of environments leads to a notable reduction in training time. Specifically, training with 16,384 environments was over 3.3$\times$ faster than with 2,048 environments. It's important to note, however, that an excessive number of environments can adversely affect the overall fitness of agents, particularly when the horizon is inadequately extended, making the RL objective overly short-term focused~\cite{makoviychuk_isaac_2021}. Consequently, we selected 8,192 parallel environments for optimal balance in subsequent experiments.

Moreover, when comparing the total duration for a full evolutionary cycle ($P=100, T=50, W=10$), DARLEI significantly outperforms DERL. In trials evolving 600 morphologies, DARLEI completed runs in approximately $205\pm8$ minutes, equating to 3.41 minutes per agent per worker. This contrasts with DERL's 16 hours for 4,000 morphologies, indicating a substantial $20.3\times$ speedup by DARLEI. Additional compute nodes can further reduce the total time dramatically.

\subsection{Impact of Simulation Parameters}

\begin{figure}[h]
    \centering
    \begin{subfigure}{0.49\textwidth}
        \centering
        \includegraphics[width=1.0\linewidth, height=7cm]{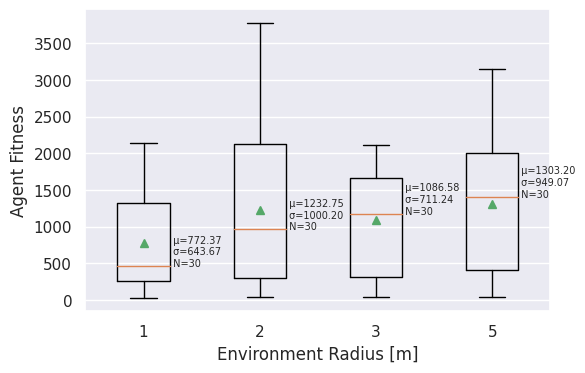} 
        \label{fig:env_radius_vs_reward}
    \end{subfigure}
    \hfill
    \begin{subfigure}{0.49\textwidth}
        \centering
        \includegraphics[width=1.0\linewidth, height=7cm]{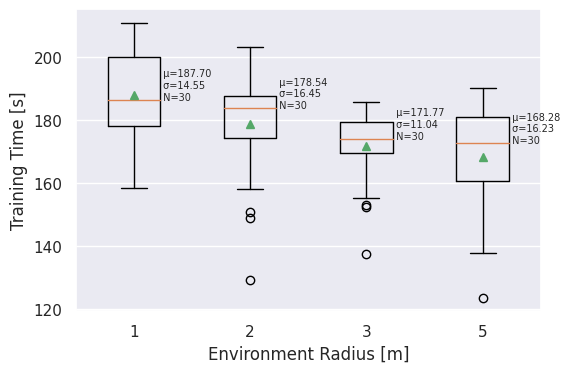}
        \label{fig:env_radius_vs_train_time}
    \end{subfigure}
    \caption{Impact of environment radius on agent fitness (left) and training time (right). Results based on 30 agents from the initial population in a simulation with 8192 parallel environments and horizon length of 16.}
    \label{fig:env_radius}
\end{figure}

\begin{figure}[h]
    \centering
    \begin{subfigure}{0.3\textwidth}
        \centering
        \includegraphics[width=1.0\linewidth, height=4cm]{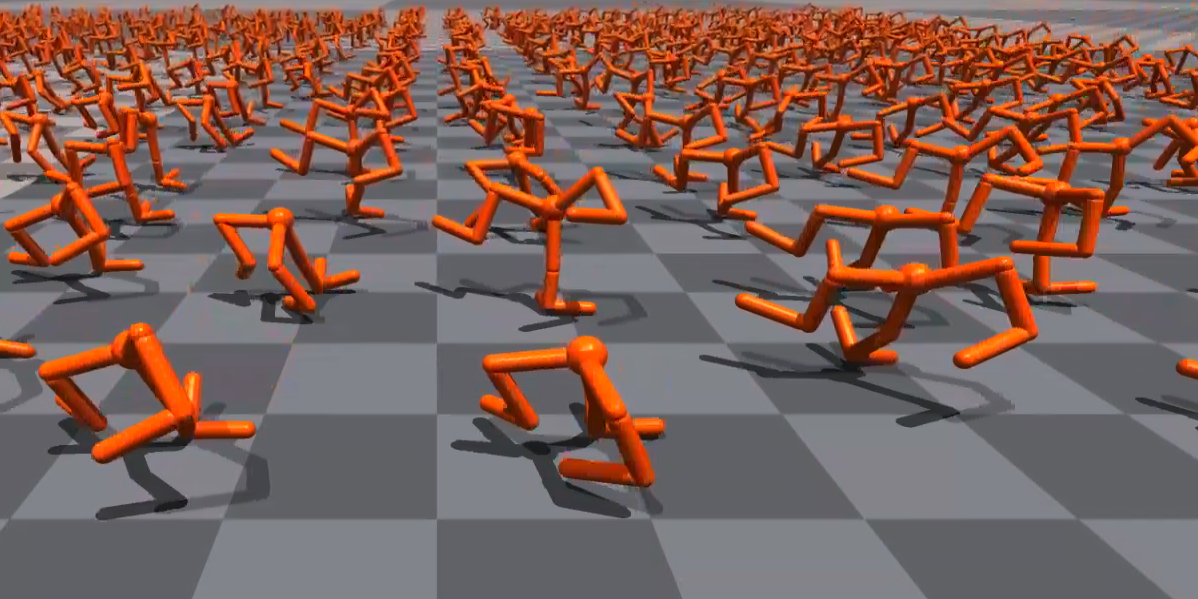} 
        \label{fig:Spacing_example_1m}
    \end{subfigure}
    \hfill
    \begin{subfigure}{0.3\textwidth}
        \centering
        \includegraphics[width=1.0\linewidth, height=4cm]{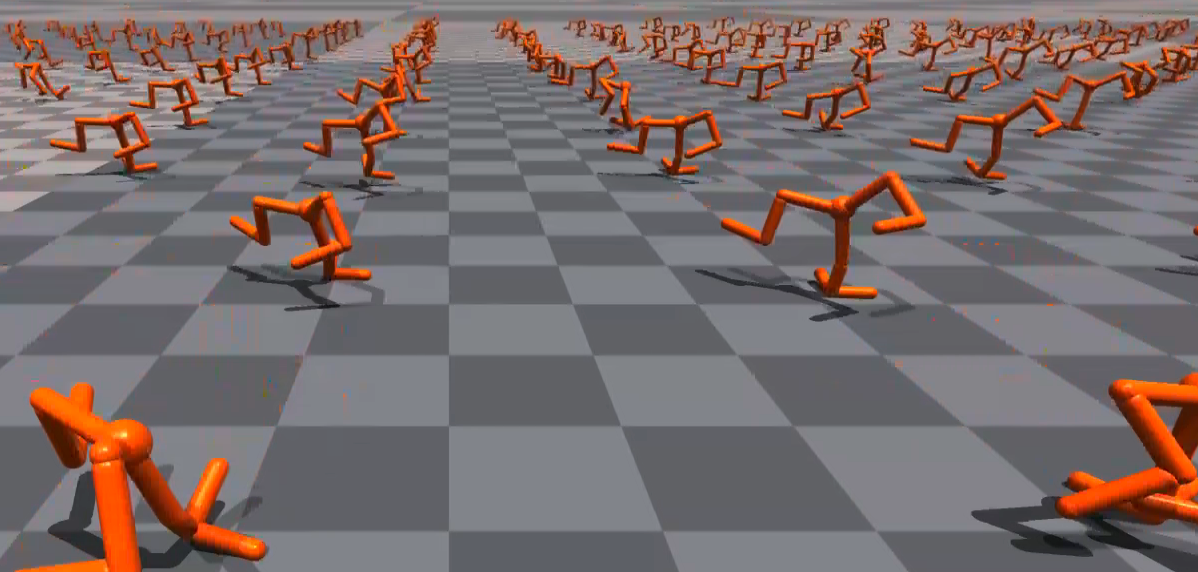}
        \label{fig:Spacing_example_2m}
    \end{subfigure}
    \hfill
    \begin{subfigure}{0.3\textwidth}
        \centering
        \includegraphics[width=1.0\linewidth, height=4cm]{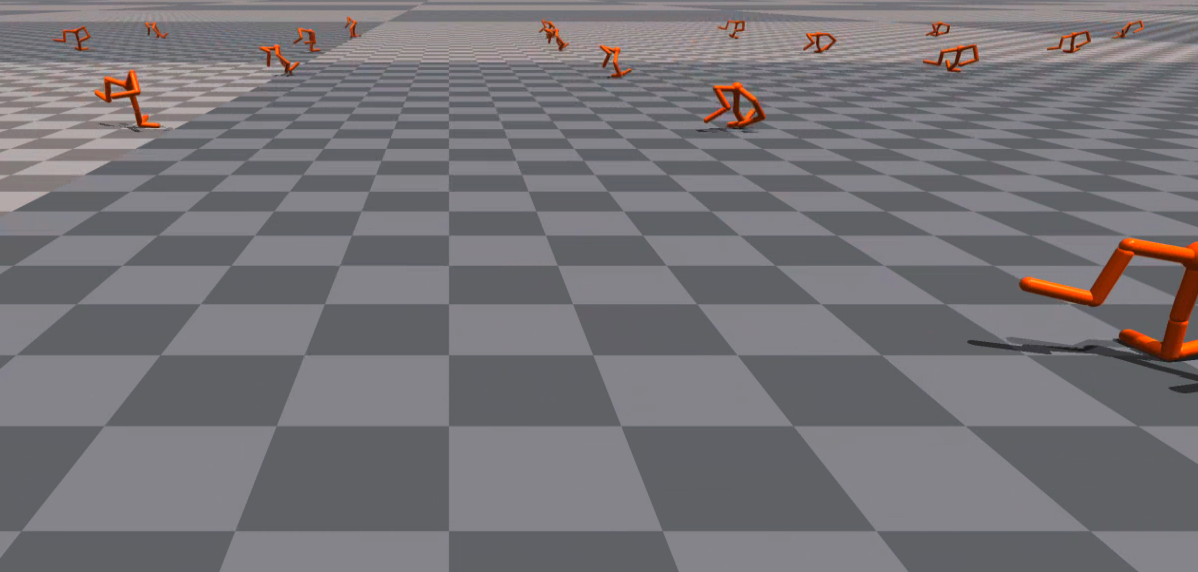}
        \label{fig:Spacing_example_5m}
    \end{subfigure}
    \caption{Environments with radii of 1m, 2m, and 5m.}
    \label{fig:spacing_examples}
\end{figure}

Investigating how varying simulation parameters affect learning, we focused on the impact of environment radius(Figure~\ref{fig:spacing_examples}). As demonstrated in Figure~\ref{fig:env_radius}, larger radii generally lead to improved median fitness, allowing agents more exploration space before encountering termination events such as loss of balance or collisions. Conversely, smaller radii induce earlier collisions, fostering the development of more robust policies. Interestingly, agents operating in a 2-meter radius displayed agile behaviors, including high-jumping and cartwheeling, suggesting that this radius acts as a 'sweet spot' for encouraging dynamic strategies. However, smaller radii also lead to increased termination frequency, prolonging training time. The optimal radius, therefore, seems to strike a balance between fostering robustness from frequent collisions and providing ample space for exploration.

\subsection{Quality of Generated Solutions}

\begin{figure}[!h]
\centering
\begin{subfigure}{\textwidth}
\includegraphics[width=1.0\linewidth, height=5.5cm]{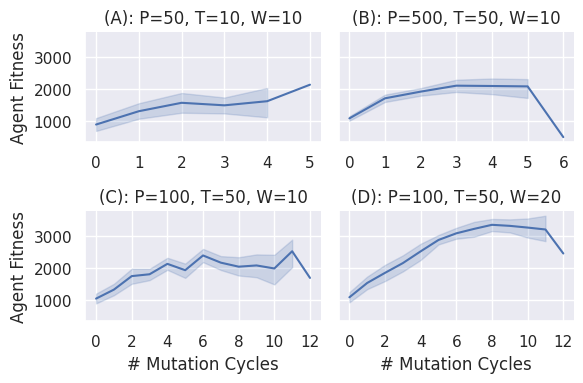} 
\label{fig:mut_cycles_vs_reward}
\end{subfigure}
\begin{subfigure}{\textwidth}
\includegraphics[width=1.0\linewidth, height=5.5cm]{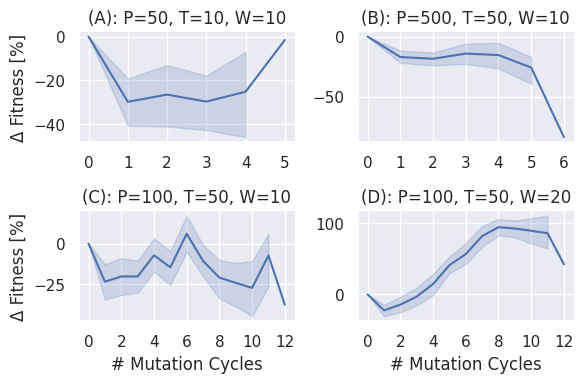}
\label{fig:mut_cycles_vs_relative_improvement}
\end{subfigure}
\begin{subfigure}{\textwidth}
\includegraphics[width=1.0\linewidth, height=5.5cm]{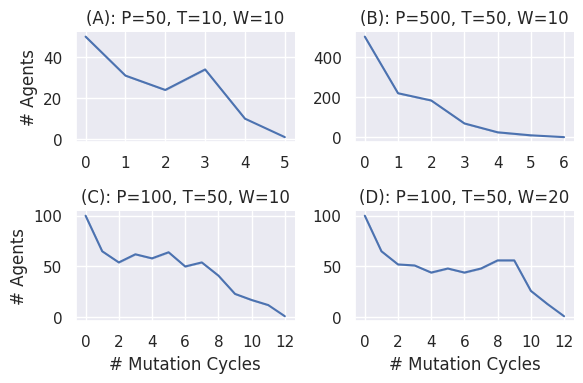}
\label{fig:mut_cycles_vs_num_agents}
\end{subfigure}
\caption{Impact of mutation cycles on: agent fitness (top), percentage improvement in fitness between the youngest child and oldest ancestor (middle), and number of agents (bottom). Results are shown for 4 experiments (A,B,C,D), each with a unique configuration of initial population size $P$, number of tournaments $T$, and number of parallel worker processes $W$.}
\label{fig:mutation_cycles}
\end{figure}

Our analysis of the evolved solutions focused on the mutation cycles, defined by the number of mutations an agent undergoes. Four experiments are conducted with varying population sizes, tournament counts, and asynchronous worker processes. Across those various experimental setups, two significant patterns emerged. First, mutations were generally harmful rather than beneficial. The top plot in Figure~\ref{fig:mutation_cycles} shows agent fitness increasing with more mutations, falsely implying mutations improve fitness. However, the center plot shows that in most experiments, mutations actually reduce fitness between ancestors and descendants on average. The fitness increase in the top plot stems from selection bias - fitter ancestors reproduce more, accumulating additional mutations.

\begin{figure}
  \centering
  \includegraphics[width=1.0\linewidth]{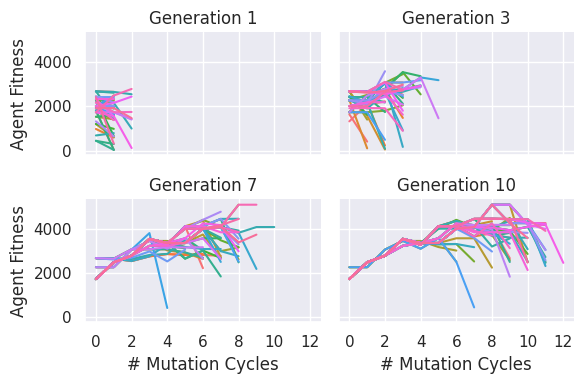}

  \caption{Population diversity is seen to decrease over generations. Lines trace rewards of agents' lineages.}
  \label{fig:gen_lack_of_diversity}
\end{figure}

Secondly, population diversity was observed to diminish rapidly across generations as shown in Figure~\ref{fig:gen_lack_of_diversity}. Notably, all final agents in the experiments traced their lineage back to just two original ancestors, despite starting from a diverse population pool. This rapid convergence highlights the need for additional mechanisms to preserve diversity and promote open-ended evolution. Potential strategies for future exploration could include speciation, fitness sharing, and the implementation of novelty search criteria to incentivize the discovery of unique strategies and behaviors.

\section{Discussion and Future Work}
Our findings underscore a critical challenge in DARLEI's framework: maintaining diversity and promoting continuous innovation in evolutionary learning systems. Notably, our experiments revealed a tendency for agent populations to converge towards humanoid-like morphologies, when tasked with traversing flat terrain. This observation may partly reflect the specific constraints of our simulated task, drawing parallels to the concept of ecological niches as discussed by Brant in minimal criterion coevolution~\cite{brant_minimal_2017}. In natural ecosystems, species evolve to fill diverse niches, each defined by unique environmental demands. Similarly, in our simulation, the \textit{niche} created by the task of flat terrain traversal seems to favour humanoid forms. While interesting, this analogy between natural ecological niches and our simulation environment could also be limited. A simpler explanation may be that the observed convergence in our model may be less about the inherent superiority of humanoid forms in flat terrain environments and more about their alignment with the specific reward function we used. This insight highlights the potential value of incorporating mechanisms akin to genetics-based speciation to enhance morphological diversity. Such mechanisms could mitigate the tendency towards convergence, enabling a broader spectrum of morphological adaptations, irrespective of task-specific advantages.

To promote greater open-endedness in future iterations of DARLEI, we could modify the fitness criteria to reward novelty and diverse problem-solving approaches over mere task performance. Approaches like Minimal Criteria Novelty Search~\cite{gomes_novelty_2014} or novelty search in coevolution~\cite{lehman_revising_2010} could be crucial in driving morphological and behavioral diversity. By valuing agents for their unique strategies in task execution and integrating both extrinsic goals and intrinsic motivations, we could potentially prevent premature convergence and cultivate a richer variety of evolved agents.

Moreover, introducing more complex, procedurally generated environments that adhere to Minimal Criterion Coevolution principles could help bridge a unique kind of \textit{sim2real} gap: not just the one in robotics, but also the divide between artificial life (\textit{alife}) worlds and evolutionary reality. These environments, if equipped with multi-objective reward systems, may enable diverse agents to succeed in various ways. The simultaneous evolution of agents and their environments could stimulate the emergence of new adaptive strategies, fostering a cycle of continuous innovation and adaptation that embodies the essence of open-ended evolution.

To validate and expand upon these findings, future iterations of DARLEI could explore a variety of tasks favoring different morphologies. This approach would help determine whether the observed convergence is task-specific or a more general characteristic of the learning and evolutionary process within our framework. Diversifying the tasks and environmental challenges will enable a more accurate assessment of the impact of task design on evolutionary outcomes and open avenues for investigating the full potential of open-ended evolution in computational models.

DARLEI introduces a new, efficient framework for decoupling immediate, individual learning from broader evolutionary processes. This separation enables rapid development and testing of various integrations of reinforcement learning with evolutionary principles. Despite current limitations in diversity, we hope to extend the framework to include elements like coevolving populations, multi-objective rewards, and novelty-driven objectives. Utilizing richer scenes at larger scales, as demonstrated by Minedojo~\cite{fan2022minedojo}, these enhancements could significantly advance our pursuit of facilitating open-ended evolution~\cite{standish2003open}. As acceleration speed and efficiency of our simulations improve, we can lower the barrier even further for research at the intersection of evolutionary dynamics, complex multi-agent interactions, and the study of embodied intelligence.

{
\small

\bibliography{references}
}

\end{document}